\documentclass[letterpaper]{article} 
\usepackage{aaai23}  
\usepackage{times}  
\usepackage{helvet}  
\usepackage{courier}  
\usepackage[hyphens]{url}  
\usepackage{graphicx} 
\usepackage{natbib}  
\usepackage{caption} 
\usepackage{algorithm}
\usepackage{algorithmic}
\usepackage{booktabs}
\usepackage{bbding}
\usepackage{makecell}
\usepackage{amsmath}
\usepackage{color}
\usepackage{soul}

\usepackage{newfloat}
\usepackage{listings}
\DeclareCaptionStyle{ruled}{labelfont=normalfont,labelsep=colon,strut=off} 
\floatstyle{ruled}
\newfloat{listing}{tb}{lst}{}
\floatname{listing}{Listing}
\title{Locate Then Generate: Bridging Vision and Language with Bounding Box for Scene-Text VQA}
\author{
    Yongxin Zhu\textsuperscript{\rm 1,3},
    Zhen Liu\textsuperscript{\rm 2,3},
    Yukang Liang\textsuperscript{\rm 2,3},
    Xin Li\textsuperscript{\rm 4},
    Hao Liu\textsuperscript{\rm 4},
    Changcun Bao\textsuperscript{\rm 4},
    Linli Xu\textsuperscript{\rm 2,3}
}
\affiliations{
    \textsuperscript{\rm 1}School of Data Science, University of Science and Technology of China\\
    \textsuperscript{\rm 2}School of Computer Science and Technology, University of Science and Technology of China\\
    \textsuperscript{\rm 3}State Key Laboratory of Cognitive Intelligence\\
    \textsuperscript{\rm 4}Tencent YouTu Lab\\
    \{zyx2016, liuzhenz, liangyukang\}@mail.ustc.edu.cn, 
    \{fujikoli, ivanhliu, changcunbao\}@tencent.com, 
    linlixu@ustc.edu.cn
}

\usepackage{bibentry}

\begin{document}

\maketitle

\begin{abstract}
In this paper, we propose a novel multi-modal framework for Scene Text Visual Question Answering~(STVQA), which requires models to read scene text in images for question answering. Apart from text or visual objects, which could exist independently, scene text naturally links text and visual modalities together by conveying linguistic semantics while being a visual object in an image simultaneously. Different to conventional STVQA models which take the linguistic semantics and visual semantics in scene text as two separate features, in this paper, we propose a paradigm of ``Locate Then Generate" (LTG), which explicitly unifies this two semantics with the spatial bounding box as a bridge connecting them. Specifically, at first, LTG locates the region in an image that may contain the answer words with an answer location module~(ALM) consisting of a region proposal network and a language refinement network, both of which can transform to each other with one-to-one mapping via the scene text bounding box. Next, given the answer words selected by ALM, LTG generates a readable answer sequence with an answer generation module~(AGM) based on a pre-trained language model. As a benefit of the explicit alignment of the visual and linguistic semantics, even without any scene text based pre-training tasks, LTG can boost the absolute accuracy by $+6.06\%$ and $+6.92\%$ on the TextVQA dataset and the ST-VQA dataset respectively, compared with a non-pre-training baseline. We further demonstrate that LTG effectively unifies visual and text modalities through the spatial bounding box connection, which is underappreciated in previous methods.
\end{abstract}

\section{Introduction}
\begin{figure}[h]
    \centering
    \includegraphics[width=0.45\textwidth]{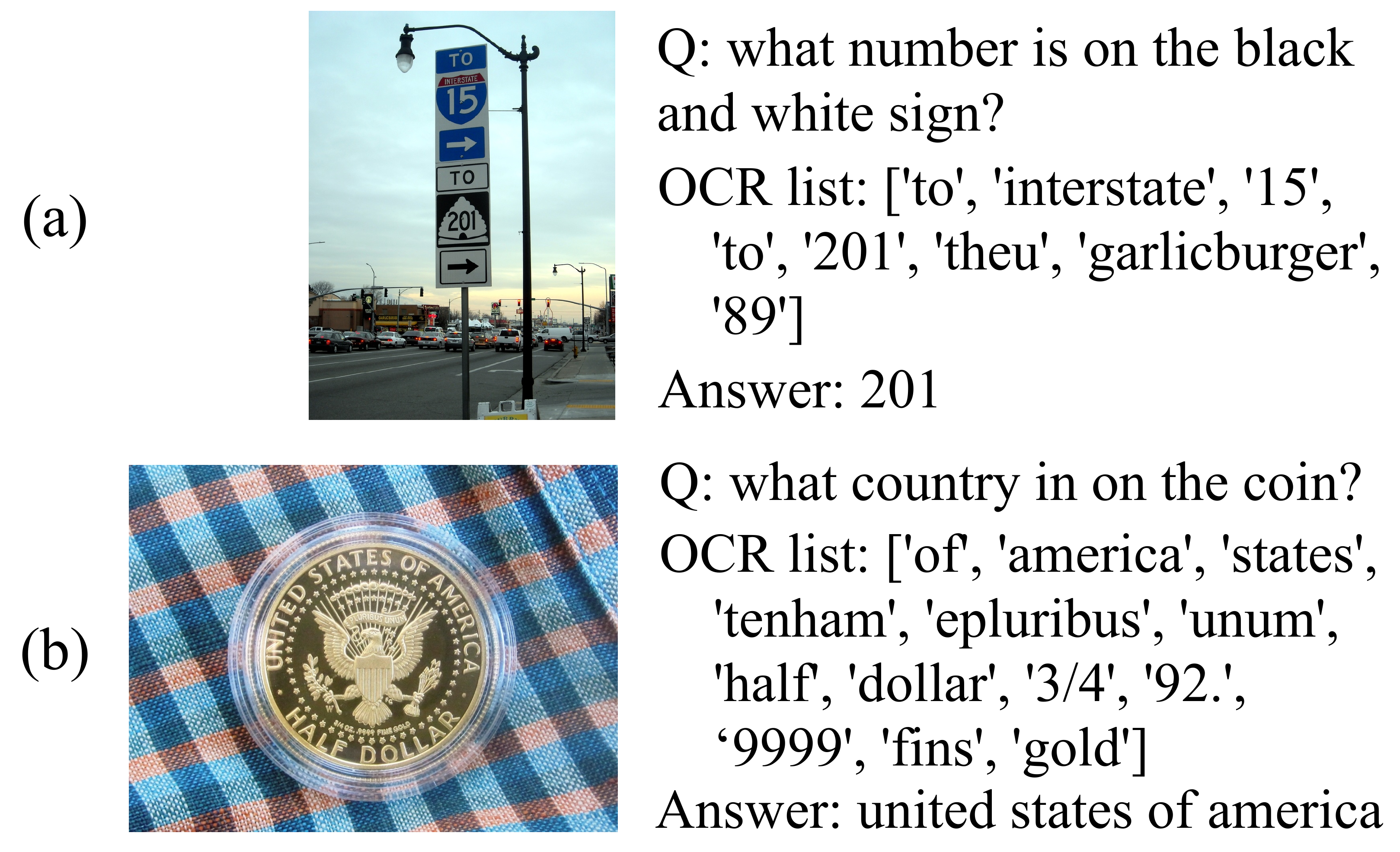}
    \caption{Examples in the TextVQA dataset~\citep{singh2019towards}. (a) The question requires models to consider linguistic semantics and visual semantics simultaneously. (b) Scene text recognition mistakes in the STVQA task. }
    \label{fig:demo_intro}
\end{figure}

The vision-language tasks that incorporate scene text, such as STVQA~\citep{singh2019towards, biten2019scene}, Text Caption~\citep{sidorov2020textcaps}, require models to reason over different modalities, including language~(question text), vision~(objects in an image), and a mixture of them~(scene text in an image which has linguistic and visual semantics). Scene text based multi-modal tasks have many potential applications, including assisting visually-impaired people~\citep{bighamVizWiz}, interaction in augmented reality~\citep{li2020mapping}, robotics~\citep{anderson2018vision} and automatic driving~\citep{9186709}, etc. Although some vision-language models~\citep{chen2019uniter, wang2021simvlm} have shown their effectiveness in learning task-agnostic vision-language joint representations, most of them focus on vision-language understanding tasks such as image-text retrieval~\citep{young2014image}, visual question answering~\citep{antol2015vqa}, visual grounding~\citep{kazemzadeh2014referitgame}, etc., which ignore scene text in images. As a consequence, they are not capable of handling scene text based VL tasks~\citep{singh2019towards}. 

Recently, several methods~\citep{hu2020iterative, yang2021tap, biten2022latr} have been proposed to augment visual-language models with the ability to read texts in images. They improve the models' performance by relying on powerful attention mechanisms and elaborately designed features of scene text. Among them, M4C~\citep{hu2020iterative} introduces the transformer~\citep{vaswani2017attention} module to fuse different modalities, followed by proposing a multi-step multi-choice decoding module to generate answer sequences step by step from the OCR list or 
frequent words in natural languages. TAP~\citep{yang2021tap} designs several scene text based pre-training tasks to implicitly learn the relationship between scene text and other modalities. Inspired by the success of LayoutLM~\citep{xu2020layoutlm}, LaTr~\citep{biten2022latr} is proposed with a layout-aware pre-training task to learn the spatial information of scene text in images. Despite the improvements achieved by the methods above in the task of STVQA, they take the linguistic semantics and visual semantics of scene text as two separate features, and rely on the self-attention mechanism to implicitly learn the relationship between them, which is hard to be further enhanced because of the lack of image-text-scene text triplet data. In addition, it is found in~\citet{zeng2021beyond} that some scene text based models~\citep{hu2020iterative} are not able to effectively understand the visual information, which also happens in the current SOTA model LaTr~\citep{biten2022latr}. 

To tackle these issues, we take a different perspective by explicitly binding the linguistic semantics and visual semantics in scene text with the location information, which takes the form of spatial bounding boxes. Specifically, we propose a ``Locate Then Generate"~(LTG) framework, which consists of an answer location module~(ALM) and an answer generation module~(AGM). Generally, we use ALM to select scene text words which may be contained in answers, and proceed to generate readable answer sequences with AGM according to the words selected by ALM. 

By locating the scene text words relevant to the answers, the answser location module (ALM) unifies the linguistic semantics and visual semantics of scene text, benefiting the comprehension of scene text from a linguistic perspective and a visual perspective simultaneously. To achieve that, we design two networks to locate the region in an image that may contain the answer words. Firstly, we use a region proposal network to roughly predict the bounding box of the answer region, which we transform to the probability space on the scene text words with one-to-one mapping between scene text words and bounding boxes. Next, we leverage a language refinement network to refine the probability predicted by the region proposal network. At last, we select the candidate answer words according to the refined probability. An example is shown in Figure~\ref{fig:demo_intro}~(a) to describe how our ALM works, in which the question requires the model to understand two concepts, including the visual semantic ``color'' and the text semantic ``number''. On the one hand, if the model is not capable of reading texts in images and identifying the textual concept ``number'', it will get confused when selecting from the black and white words ``to'' and ``201''; on the other hand, if the model can not understand the visual information regarding ``black and white", it will get confused in selecting numbers ``15'' or ``201''. In our ALM, the region proposal network is responsible for understanding the visual semantics like ``color'', and the language refinement network is responsible for understanding the text semantics like ``number'' mentioned above. By unifying them through the bounding box mapping, ALM is enabled to benefit from both of them.

For the answer generation module~(AGM), we leverage a pre-trained language model BART~\citep{lewis2019bart} to generate readable answer sequences auto-regressively with the words selected by the answer location module. Compared with the previous methods that can only generate words through an amalgamation of a pointer mechanism and a 5K most frequent vocabulary which is data-specific~\citep{hu2020iterative, yang2021tap}, our AGM can generate answer words out of OCR lists, which is also adopted in LaTr~\citep{biten2022latr}. As well known, scene text of arbitrary shape is difficult to be recognized~
\citep{DBLP:conf/cvpr/LiuCSHJW20}. For example in Figure~\ref{fig:demo_intro}(b), the OCR system~(Microsoft OCR system) recognizes the key scene text words ``states'', ``of'',``america'', while missing the word ``united''. In previous methods~\citep{hu2020iterative, yang2021tap}, the model can only generate answers like ``states of america'' because of the fixed OCR vocabulary. In contrast, our model can generate ``united states of america'' for the outstanding denoising ability of pre-trained language models. 

We conduct experiments with the proposed LTG framework on the TextVQA and ST-VQA datasets. LTG improves the accuracy on the TextVQA dataset from $44.50\%$ to $50.56\%$, compared with a non-pre-training baseline. Moreover, our final model ranks \textbf{No.2}\footnote{According to the official leader-boards~(August. 2022). Note that the No.1 model GIT 
is pretrained on an incredibly huge dataset with 0.6B image-text pairs.} on the ST-VQA challenge, and outperforms the best non-pretraning baseline on the ST-VQA dataset by $+6.92\%$ in absolute accuracy, even outperforming the pre-trained model TAP~\citep{yang2021tap} by $+1.83\%$ in absolute accuracy.

In summary, the main contributions of the work include: 
\begin{itemize}
\item \noindent We propose a novel and effective framework of ``Locate Then Generate" (LTG) to explicitly leverage the relationship between the text words and spatial bounding boxes in scene text, unifying the linguistic semantics and visual semantics of scene text, leading to significant performance improvement on the STVQA task.
\item \noindent We propose to exploit a pre-trained denoising language model for answer generation, which can correct some OCR recognition errors effectively.
\end{itemize}

\section{Related Work}
\noindent \textbf{Vision-Language Tasks Incorporating Scene Text. } Recently, some multi-modal models~\citep{chen2019uniter, wang2021simvlm} have achieved outstanding performance in vision-language tasks like VQA~\citep{antol2015vqa}, image caption~\citep{anderson2018bottom}, image-text retrieval~\citep{young2014image}, visual grounding~\citep{kazemzadeh2014referitgame}, etc. However, recent studies~\citep{singh2019towards} show that these models fail to read text in images. To address this problem, some methods have been proposed to augment vision-language models with the ability to read scene text in natural images.
Among them, LoRRA~\citep{singh2019towards} is the first model that is able to read scene text with an OCR brach based on a VQA model Pythia~\citep{jiang2018pythia}. In M4C~\citep{hu2020iterative} a transformer~\citep{vaswani2017attention} module is introduced followed by a multi-step multi-choice decoding module to generate answer sequences, which becomes the backbone of many subsequent models. For example, SA-M4C~\citep{kant2020spatially} extends M4C by providing supervision on self-attention weights. In MM-GNN~\citep{gao2020multi}, a representation of three graphs is proposed for three modalities, with three aggregators to update the message passing. Instead of designing separate graphs for each modality, SMA~\citep{gao2021structured} encodes all modalities into one single graph. SSBaseline~\citep{zhu2021simple} splits OCR token features into separate visual and linguistic parts, which are fused pair-wisely before being sent to a transformer decoder to generate answers. LOGOS~\citep{lu2021localize} extracts ROI features to align question semantics and visual semantics, followed by a scene text clustering operation to enhance the spatial information. TAP~\citep{yang2021tap} proposes to pre-train the model on a large image-text-scene text triplet dataset OCR-CC~(1.4M image-text-scene text triplets) and designs three auxiliary tasks, including masked language modeling~(MLM), image-text matching~(ITM) and relative position prediction~(RPP) to enhance its ability to capture the contextualized information of scene text. More recently, LaTr~\citep{biten2022latr} proposes a layout-aware multi-modal pre-training task based on T5~\citep{raffel2020exploring} with an extremely large Industrial Document Library~(64M pages of document images). Additionally, some big multi-modal pre-trained models~\citep{Alayrac2022FlamingoAV, wang2022git} also perform tests on the STVQA task and 
achieve competitive results.

Despite the powerful transformer module, most previous works directly fuse the vision features, linguistic features and spatial features together into the transformer block. Such a rough fusion design could be ineffective in learning the aligned multi-modal representations and thus limit the model performance. In this study, we explicitly leverage the one-to-one map between text words and spatial locations in scene text, with which we bridge the vision and language semantics.

\noindent \textbf{Visual Grounding. }
Visual grounding aims to predict the location of a region referred by the language expression in an image. Recent advances in visual grounding can be categorized into two-stage methods~\citep{YuMAttNet, ZhangNC18} and one-stage methods~\citep{DBLP:conf/iccv/YangGWHYL19, DBLP:conf/iccv/DengYCZL21, KamathSLSMC21}. Two-stage methods usually generate region proposals in the first stage by a pre-trained object detector~\citep{YuMAttNet, ZhangNC18} followed by leveraging the language expressions to select the best matching region in the second stage. In comparison, instead of keeping the computation-intensive region proposal generation in the two-stage paradigm, one-stage methods employ a multi-modal network to densely fuse different modalities and then predict the bounding box in one step. 
Recent works mostly rely on transformer to learn the relationship between the text modality and visual modality. For example, TransVG~\citep{DBLP:conf/iccv/DengYCZL21} uses a transformer based encoder-decoder architecture to directly regress the object bounding box.
MDETR~\citep{KamathSLSMC21} builds a modulated end-to-end detector with a transformer-based architecture to reason jointly over texts and images by fusing the two modalities at an early stage, followed by a non-auto-regressive transformer decoder to locate the objects referred to in texts.

\section{Method}
In this section, we elaborate on the proposed LTG framework for the STVQA task in detail. We start with the problem definition, followed by the motivation and the model architecture. We then proceed to introduce how the model is designed.

\subsection{Problem Definition}
Given an image and a question text in the STVQA task, the image usually contains many scene text objects represented as text words and bounding boxes, which are aligned with one-to-one mapping. The task of STVQA requires the model to generate an answer text to the question, which is usually a sequence of words either from the scene text word list or the dictionary of the pre-trained language model. 

\begin{figure*}[t]
    \centering
    \includegraphics[width=0.6\textwidth]{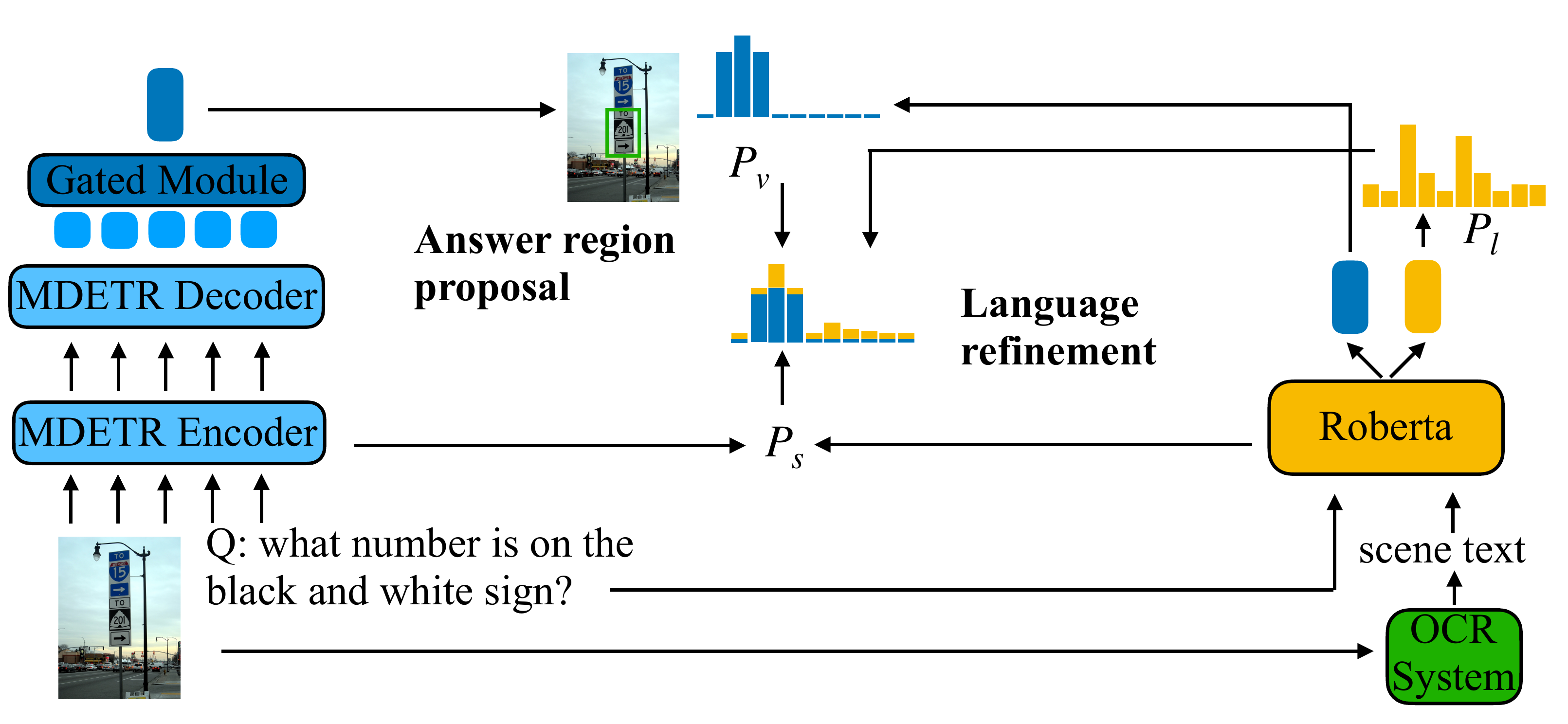}
    \caption{An illustration of the answer location module in LTG. It consists of an answer region proposal network and a language refinement network. The answer region proposal network predicts the region bounding box which may contain the answer words from a visual perspective. The language refinement network refines the probability space from a language perspective.}
    \label{fig:model_pic}
\end{figure*}

\subsection{Model Architecture}
We aim to build a framework that is able to explicitly leverage the relationship between the visual semantics and 
linguistic semantics of the scene text. To be specific, the model should understand what the scene text words mean in linguistics, and what visual attributes they have in images simultaneously. To achieve this goal, we propose a novel ``Locate Then Generate"~(LTG) framework with an answer location module (ALM) and an answer generation module (AGM). 

Specifically, ALM aims to understand the visual information of the image, and then locate the region that may contain the words appearing in the answer text. Different from previous STVQA models that directly generate answer text sequences according to the image-question pair and the scene text 
word list, which implies that the model may only use the text information to answer the question instead of unifying the text and vision modalities together, we design an answer location module to force the model to look at the image. An illustration of ALM is shown in Figure~\ref{fig:model_pic}, which consists of an answer region proposal network and a language refinement network. The former network is based on a visual grounding model to predict the bounding box of the answer region visually, which sometimes may include ambiguous scene text words sharing the same visual attributes with the correct scene text words in images or appearing close to them. In order to select or filter out the correct scene text words, we design a language refinement network based on a pre-trained language model to distinguish whether they are correct or not from the linguistic perspective. Thus, the model can select scene text words by unifying the linguistic semantics and visual semantics simultaneously. Next, AGM takes the scene text words selected by ALM as input, which are usually some disordered and sometimes wrong words, to generate ordered readable correct answer sequences. We introduce the details as follows.

\subsection{Pre-processing}
We first explain how to construct answer region box targets from the answer sequences. For an answer sequence $s_a=[w^a_1,w^a_2,\dots,w^a_n]$, a scene text word list $s_o=[w^o_1,w^o_2,\dots,w^o_m]$ and its associated bounding boxes $b_o=[b^o_1,b^o_2,\dots,b^o_m]$, where the tokens in $s_a$ are either from $s_o$ or the dictionary of the pre-trained language model. Each bounding box $b^o_i=[x_1^i,y_1^i,x_2^i,y_2^i]$ in $b_o$ indicates the relative positions in images, where $[x_1^i,y_1^i]$ corresponds to the
position of the upper left corner of the bounding box, while $[x_2^i,y_2^i]$ represents the position of the lower right corner. To construct the answer region boxes, we will conduct an exact matching between the answer sequence $s_a$ and the scene text word list $s_o$ word by word. Suppose answer words $[w^a_i,w^a_j]$ in $s_a$ match the scene text words $[w^o_k,w^o_l]$ in $s_o$, and the corresponding bounding boxes are $b^o_k=[x^k_1,y^k_1,x^k_2,y^k_2]$ and $b^o_l = [x^l_1,y^l_1,x^l_2,y^l_2]$, we set the answer region box as below:

\begin{equation}
\begin{split}
b_a=&[\min (x^k_1,x^l_1),\min (y^k_1,y^l_1),\\
&\max (x^k_2,x^l_2),\max (y^k_2,y^l_2)]
\end{split}
\end{equation}

If no words are matched, we simply set the answer region box as $[x_1=0,y_1=0,x_2=0,y_2=0]$. In the language refinement network, we 
formulate the task of language refinement as a binary classification 
problem on the scene text word list $s_o$. For each word $s^o_i$ in $s_o$, we set its label tag in $y\in \{0,1\}$, where $y=1$ indicates that it appears in $s_a$ and $y=0$ otherwise. In the above example, $w^o_k,w^o_l$ are tagged as $y_k=1,y_l=1$, while the rest are tagged as $y_{i\neq k,l}=0$.

\subsection{Answer Location Module}
We build our answer location module based on MDETR~\citep{KamathSLSMC21}, a pre-trained end-to-end modulated detector that detects objects in an image conditioned on a raw text query. Following the original input of MDETR, we encode an image by a convolution backbone and flatten it. In order to conserve the spatial information, we add 2-D position embeddings to this flattened vector. The question text is encoded by a pre-trained transformer language model Roberta~\citep{liu2019roberta} to produce a sequence of hidden vectors. Then the concatenated feature vectors are fed into a multi-modal transformer encoder-decoder to generate the bounding boxes, with a $[\text{CLS}]$ special token prepended. In order to fully exploit the bounding box information, we use a layout enhanced Roberta model~\citep{wang2022lilt} instead of the vallina Roberta, which can output the original language hidden states and layout hidden states separately.

To roughly predict the bounding box of the answer region, we design a region proposal network. On one side, we concatenate the language hidden states $h_{lang}$ and the layout hidden states $h_{lay}$ of the layout enhanced Roberta to predict the probability $P_l$ regarding which token could be the answer from the linguistic perspective,
\begin{equation}
\begin{split}
& h_l = \text{concat}(h_{lang}, h_{lay}), \\
& P_l = \sigma (w_l h_l + b_l),
\end{split}
\end{equation}
where $w_l$, $b_l$ are learnable parameters, and $\sigma$ is the sigmoid function. 

On the other side, we apply a gated module on the hidden states of the MDETR decoder $h_v$ to get the visual features of the answer region $h^v_a$,

\begin{equation}
h^v_a = \sum \sigma (w_v h_v + b_v) h_v
\end{equation}

In order to enhance the region proposal network's capability of spatial perception, we aggregate the layout hidden states $h_{lay}$ with a gated module as the spatial features $h^s_a$ to augment the visual features $h^v_a$ of the answer region. With both visual features and spatial features of the answer region, we concatenate them together as the final representation of the answer region proposal $h_a$. Then we apply a bounding box regression module with a two-layer feed-forward network to predict the bounding box of the answer region $b_p$,

\begin{equation}
\begin{split}
& h^s_a = \sum P_l h_{lay}, \\
& h_a = \text{concat}(h^v_a, h^s_a), \\
& b_p = \sigma (\text{FFN}_{bbox}(h_a))
\end{split}
\end{equation}

The loss function of answer region regression is the same as in MDETR:
$$
\mathrm{Loss}_{bbox} = \lambda_1 L_1(b_p, b_a) + \lambda_2 \text{GIOU}(b_p, b_a),
$$
where $L_1$ indicates the L1 distance between boxes and GIOU is the generalized IOU distance~\citep{Rezatofighi_2018_CVPR}.

Notice that each scene text word is associated with a bounding box, we can define the proportion of the overlapping area between bounding boxes as the probability in the language space of scene text. Thus the proportion of the overlapping area between the bounding box of the predicted answer region 
and that of each scene text word can be seen as the probability $P_v$ regarding which token could be the answer from the visual perspective. We design a modified $\hat{\mathrm{IOU}}$ metric to measure the proportion of the overlapping area:
\begin{equation}
P_v = \hat{\mathrm{IOU}} = \frac{|A\cap B|}{|B|},
\end{equation}
where $A$ is the predicted bounding box of the answer region, $B$ is the bounding box of the scene text word and $|.|$ indicates the area of a box. 

In order for the model to balance the probability of selecting answer words from a linguistic perspective and a visual perspective, we design a language refinement network. A selection probability $P_s\in [0,1]$ is calculated from the hidden state of the visual $[\text{CLS}]$ token $h^v_{cls}$ and the hidden state of the language $[\text{CLS}]$ token $h^t_{cls}$:

\begin{equation}
p_s = \sigma (\text{FFN}_{cls} \left(w^t_{cls} h^t_{cls} + w^v_{cls} h^v_{cls}\right))
\end{equation}

Next, $p_s$ is used as a soft switch to choose between selecting a token from the visual perspective by sampling from $P_v$, or selecting a token from the linguistic perspective by sampling from $P_l$. We obtain the following probability distribution on each scene text word $P_w$, and the loss function of word selection is the binary negative log-likelihood,

$$
P_w = p_s P_v + (1-p_s) P_l, 
$$
$$
\mathrm{Loss}_s = - \sum_{i} y_i \log p(w_i) + (1 - y_i) \log (1 - p(w_i))
$$

The ability of producing answer words from the visual spatial perspective is one of the primary advantages of our answer location module. By contrast, previous models~\citep{hu2020iterative, yang2021tap} take object features extracted by Faster-RCNN~\citep{DBLP:journals/pami/RenHG017} as vision tokens in transformer, neglecting the relationship between scene text words and scene text bounding boxes.

\begin{table*}[t]
  \centering
  \begin{tabular}{ll|ccc}
    \toprule
    Method & Extra-Training Data & Val Acc. & Val ANLS & Test ANLS \\
    \midrule
    M4C~\citep{hu2020iterative} & - & 38.05 & 0.472 & 0.462 \\
    SA-M4C~\citep{kant2020spatially} & - & 42.23 & 0.512 & 0.504 \\
    SMA~\citep{gao2021structured} & - & - & - & 0.466 \\
    CRN~\citep{liu2020cascade} & - & - & - & 0.483 \\
    LaAP-Net~\citep{han2020finding} & - & 39.74 & 0.497 & 0.485 \\
    SSBaseline~\citep{zhu2021simple} & - & - & - & 0.509 \\
    LOGOS~\citep{lu2021localize} & - & 44.10 & 0.535 & 0.522 \\
    TAP~\citep{yang2021tap} & - & 45.29 & 0.551 & 0.543 \\
    LTG~(Ours) & - & \textbf{52.21} & \textbf{0.619} & \textbf{0.598} \\
    \midrule
    $\text{SSBaseline}^{+}$~\citep{zhu2021simple} & TextVQA~(28K) & - & - & 0.550 \\
    $\text{LOGOS}^{+}$~\citep{lu2021localize} & TextVQA~(28K) & 48.63 & 0.581 & 0.579 \\
    $\text{TAP}^{+}$~\citep{yang2021tap} & OCR-CC~(1.4M) & 50.83 & 0.598 & 0.597 \\
    $\text{LTG}^{+}$~(Ours) & TextVQA~(28K) & \underline{52.70} & \underline{0.622} & \underline{0.609} \\
    LaTr~\citep{biten2022latr} & IDL~(64M) & 61.64 & 0.702 & 0.696 \\
    \bottomrule
  \end{tabular}
\caption{Results on the ST-VQA dataset~\citep{biten2019scene}. The top part of the table presents results without extra data but only the ST-VQA dataset for training, and the bottom part uses extra training datasets. Among them, ``$\text{SSBaseline}^{+}$'', ``$\text{LOGOS}^{+}$'' and our ``$\text{LTG}^{+}$'' uses the TextVQA dataset as the extra traning data, ``$\text{TAP}^{+}$'' and ``LaTr'' uses much larger datasets which are OCR-CC and IDL respectively. 
}
\label{tab:stvqa_res}
\end{table*}

Finally, the loss function of the answer selection module is the combination of the answer region regression loss and the answer word selection loss:
\begin{equation}
\mathrm{Loss}_a = \mathrm{Loss}_{bbox} + \mathrm{Loss}_s
\end{equation}

\subsection{Answer Generation Module }
To transform the words selected by ALM into answer sequences, we use the pre-trained denoising language model BART~\citep{lewis2019bart} for its excellent generation and denoising performance in NLG tasks. 

Given the question texts $s_q$, the scene text word list $s_o$ and the 
words selected by ALM $s_l$, we concatenate them into the sequence $[s_q;s_l;s_o]$ which is then fed into the BART encoder for fusion. The encoder can model the language semantic relationship between $s_l$ and $s_o$ with $s_l$ as the guidance for answer generation. The decoder is responsible for generating an answer sequence $s_a$ in an auto-regressive manner with the cross-entropy loss:
\begin{equation}
\mathrm{Loss}_g = -\sum_i \log P(s_{a;i=n}|s_q,s_l,s_o,s_{a;i<n})
\end{equation}

\begin{table*}[t]
  \centering
  \begin{tabular}{llll|cc}
    \toprule
    Method & OCR System & Pre-Training Data & Extra Finetune & Val Acc. & Test Acc.\\
    \midrule
    M4C~\citep{hu2020iterative} & Rosetta-en & \XSolidBrush & \XSolidBrush & 39.40 & 39.01 \\
    SMA~\citep{gao2021structured} & Rosetta-en & \XSolidBrush & \XSolidBrush & 40.05 & 40.66  \\
    CRN~\citep{liu2020cascade} & Rosetta-en & \XSolidBrush & \XSolidBrush & 40.39 & 40.96  \\
    LaAP-Net~(Han et al. 2020) & Rosetta-en & \XSolidBrush & \XSolidBrush & 40.68 & 40.54  \\
    SSBaseline~\citep{zhu2021simple} & SBD-Trans OCR & \XSolidBrush & \XSolidBrush & 43.95 & 44.72  \\
    $\text{M4C}^{+}$~\citep{yang2021tap} & Microsoft-OCR & \XSolidBrush & \XSolidBrush & 44.50 & -  \\
    $\text{M4C}^{+}$~\citep{biten2022latr} & Amazon-OCR & \XSolidBrush & \XSolidBrush & 47.84 & -  \\
    TAP~\citep{yang2021tap} & Microsoft-OCR & TextVQA & \XSolidBrush & 49.91 & 49.71  \\
    LOGOS~\citep{lu2021localize} & Microsoft + Rosetta & \XSolidBrush & \XSolidBrush & 50.79 & 50.65  \\
    LTG~(Ours) & Microsoft-OCR & \XSolidBrush & \XSolidBrush & 50.56 & 50.04 \\
    \midrule
    $\text{M4C}^{+}$~\citep{yang2021tap} & Microsoft-OCR & \XSolidBrush & ST-VQA & 45.22 & -  \\
    SA-M4C~\citep{kant2020spatially} & Google-OCR & \XSolidBrush & ST-VQA & 45.4 & 44.6  \\
    SMA~\citep{gao2021structured} & SBD-Trans OCR & \XSolidBrush & ST-VQA & - & 45.51  \\
    SSBaseline~\citep{zhu2021simple} & SBD-Trans OCR & \XSolidBrush & ST-VQA & 45.53 & 45.66  \\
    TAP~\citep{yang2021tap} & Microsoft-OCR & TextVQA, ST-VQA & ST-VQA & 50.57 & 50.71  \\
    LOGOS~\citep{lu2021localize} & Microsoft + Rosetta & \XSolidBrush & ST-VQA & 51.53 & 51.08  \\
    LTG~(Ours) & Microsoft-OCR & \XSolidBrush & ST-VQA & 51.04 & 50.3 \\
    \midrule
    $\text{TAP}^{+}$~\citep{yang2021tap} & Microsoft-OCR & \makecell[l]{TextVQA, ST-VQA, \\ TextCaps, OCR-CC} & ST-VQA & 54.71 & 53.97  \\
    LaTr~\citep{biten2022latr} & Amazon-OCR & IDL & ST-VQA & 61.05 & 61.60  \\
    \bottomrule
  \end{tabular}
\caption{Results on the TextVQA dataset~\citep{singh2019towards}. As commonly done, the top part of the table presents results without extra data but TextVQA dataset for training, the middle part uses the ST-VQA dataset for extra 
finetuning, and the bottom part uses extra pre-training data. Different OCR detector are listed in the ``OCR system'' column. The method ``$\text{M4C}^{+}$'' uses different OCR systems compared with ``M4C''. ``$\text{TAP}^{+}$'' uses OCR-CC for pre-training compared with ``TAP''. 
}
\label{tab:textvqa_res}
\end{table*}

\section{Experiments}
We evaluate our LTG framework on the ST-VQA~\citep{biten2019scene} and TextVQA~\citep{singh2019towards} datasets. We first briefly introduce the datasets, followed by the results and discussions.

\subsection{Datasets}
\noindent \textbf{ST-VQA\footnote{We use ST-VQA 
to denote the dataset proposed in~\citep{biten2019scene}, and STVQA
to denote the general task of scene text VQA.}.}
The ST-VQA dataset~\citep{biten2019scene} contains 21,892 images from multiple sources including 
ICDAR~\citep{karatzas2015icdar}, 
VizWiz~\citep{gurari2018vizwiz}, 
Visual Genome~\citep{krishna2017visual}, 
COCO-Text~\citep{veit2016coco}, etc. We follow the settings in previous works~\citep{hu2020iterative, yang2021tap} and split the dataset into train, validation and test splits with $17,028$, $1,893$, and $2,971$ images respectively. The methods are evaluated by both accuracy and Average Normalized Levenshtein Similarity (ANLS).

\noindent \textbf{TextVQA. } 
The TextVQA dataset~\citep{singh2019towards} contains $28,408$ images from the Open Images dataset~\citep{kuznetsova2020open}, with human-written questions asking to reason about the text in images. We follow the same training/validation/test split used in the previous work~\citep{singh2019towards} in our experiments. Similar to the VQA~\citep{goyal2017making} dataset, each question in the TextVQA dataset has 10 human annotated answers, and the final accuracy is measured via soft voting of the 10 answers.

\subsection{Results}
\noindent \textbf{ST-VQA.}
Table~\ref{tab:stvqa_res} shows the results on the ST-VQA dataset~\citep{biten2019scene}. We use the Microsoft-OCR system to extract scene text words in images and then train LTG on the ST-VQA training set. It is noteworthy that the current highest scores from TAP~\citep{yang2021tap} and LaTr~\citep{biten2022latr} are achieved by pre-training on other large-scale OCR datasets like OCR-CC~(1.4M) and IDL~(64M) which are designed to better utilize the scene text features in multi-modal tasks. 
Nevertheless, these datasets are hard to obtain. 
As a matter of fact, our proposed model is focused on fully exploiting the connections between modalities in the model design, rather than the large-scale pre-training paradigm. In the meantime, our model is fully compatible with the baselines pre-trained on these datasets. We expect a further improvement of our model when it can get access to more OCR data. 

\begin{figure}[t]
    \centering
    \includegraphics[width=0.5\textwidth]{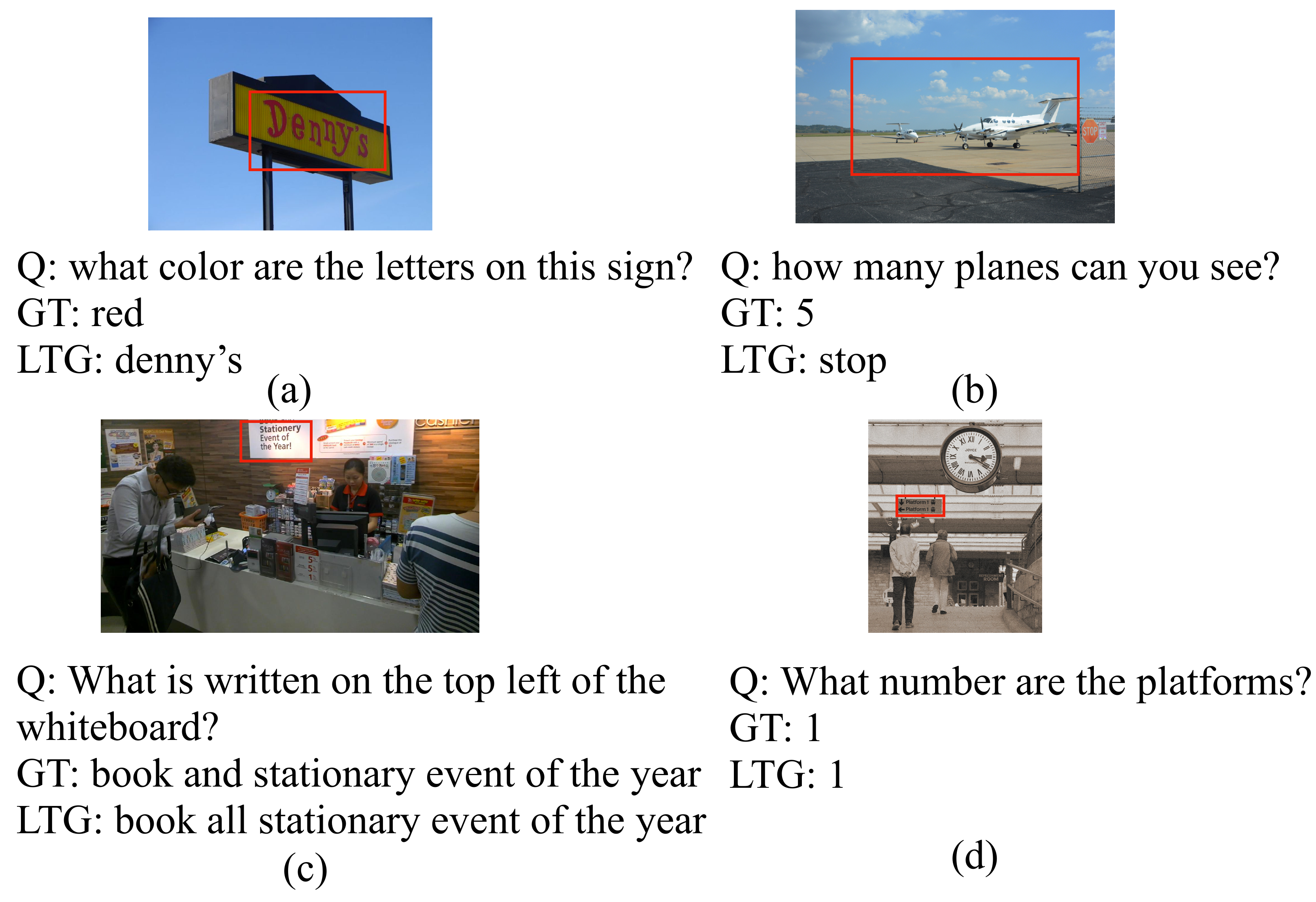}
    \caption{Examples from the TextVQA validation set and the ST-VQA validation set. The red bounding boxes are predicted by ALM in LTG.}
    \label{fig:case_study}
\end{figure}

We have several interesting observations from Table~\ref{tab:stvqa_res}: \textbf{(1)} LTG outperforms the previous non-pre-training SOTA model TAP~\citep{yang2021tap} by $+6.92\%$ and $+6.8\%$ on accuracy and ANLS respectively in the validation set, and $+5.5\%$ by ANLS in the online test\footnote{\url{https://rrc.cvc.uab.es/?ch=11&com=evaluation&task=3}}. \textbf{(2)} Surprisingly, we can see that LTG without extra training data even outperforms the pre-trained model $\text{TAP}^{+}$ by $+1.38\%$ and $+2.1\%$ on accuracy and ANLS respectively. \textbf{(3)} By joint training with the extra dataset TextVQA, which is much smaller than OCR-CC and IDL though, $\text{LTG}^{+}$ can get a further improvement in accuracy and ANLS by $+0.49\%$ and $+0.3\%$ in the validation set, $+1.1\%$ in the online test. The improvement validates the effectiveness of our method for the STVQA task.

\begin{table}[t]
  \centering
  \begin{tabular}{lccc}
    \toprule
    Model & V & L & Acc. \\
    \midrule
    (a) & \XSolidBrush & \XSolidBrush & 48.02 \\
    (b) & \Checkmark & \XSolidBrush & 50.52 \\
    (c) & \XSolidBrush & \Checkmark & 50.48 \\
    (d) & \Checkmark & \Checkmark & 52.21 \\
    \bottomrule
  \end{tabular}
\caption{Ablation Studies on the ST-VQA dataset. We ablate LTG by varying the word selection proposals of our method. V refers to selecting words according to the probability of the region proposal network in ALM, and L refers to 
selecting words according to the probability 
output by Roberta. 
Unifying both V and L refers to selecting words by ALM.}
\label{tab:ablation}
\end{table}

\noindent \textbf{TextVQA. } Table~\ref{tab:textvqa_res} reports the accuracy on the TextVQA dataset~\citep{singh2019towards}, from which two major observations can be made: \textbf{(1)} We can see significant discrepancies in the accuracy of the same model with different OCR systems, e.g., M4C with Rosetta~($39.40\%$), Microsoft-OCR~($44.50\%$), Amazon-OCR~($47.84\%$) respectively. The discrepancies indicate that the performances on the TextVQA dataset are severely limited by the quality of the scene text words detected. Thus, decoding with the fixed OCR vocabulary~\citep{hu2020iterative, yang2021tap} is not a proper method for the STVQA task. In contrast, the AGM module we propose is based on a pre-trained denoising language model, which is robust with the OCR recognition errors. \textbf{(2)} LTG outperforms the baseline model M4C by $+6.06\%$, but is slightly lower than the non-pre-training SOTA LOGOS~\citep{lu2021localize}. This may be due to the fact that the TextVQA dataset contains some questions that can be answered \textbf{without} scene text, which are similar to the questions in the VQA task~\citep{antol2015vqa}. For example in Figure~\ref{fig:case_study}(a,b), the questions require the model to answer the ``color'' and the ``number'', which 
are not consistent with the definition and assumptions of the STVQA task. However, LTG is designed for the questions of which the answers contain scene text words. 
Even though it can predict the correct answer regions for the questions, there are no scene text words for answers in them, which inevitably impairs its performance. In contrast to the TextVQA dataset, answers in the ST-VQA dataset  are more consistent with the STVQA task, 
where LTG can get a huge improvement.

\subsection{Ablation Studies}
We conduct ablation studies on the ST-VQA dataset to examine the effectiveness of our LTG framework for the STVQA task. Results are shown in Table~\ref{tab:ablation}. Row (a) 
correspond to the original BART model trained on the ST-VQA dataset by simply concatenating question and scene text words which are fed to  
the model. We can see even without any extra features or pre-training tasks, BART reaches a performance of $48.02\%$, $+2.73\%$ higher than the 
best non-pre-training baseline TAP ($45.29\%$), which proves the superiority of the pre-trained language model used in our AGM module. In rows (b)-(c), we add the 
selected words by the region proposal network and the language refinement network in ALM respectively. The accuracy is improved by $+2.5\%$ and $+2.46\%$, proving the importance of visual information in the STVQA task. After unifying them together, we can further improve the accuracy from $48.02\%$ to $52.21\%$, 
validating the effectiveness of linking language and vision through bounding 
boxes in LTG.

\subsection{Case Study and Discussion}
We conduct case studies to intuitively demonstrate the advantages of the proposed LTG framework in Figure~\ref{fig:case_study}(c)-(d). The cases are picked from the ST-VQA validation dataset.

\begin{itemize}
\item \noindent The example in Figure~\ref{fig:case_study}(c) shows the outstanding visual understanding ability of LTG. The question requires the model to understand the visual attributes ``position'' and ``color''. LTG can easily locate the corresponding region which contains the answer words from the question.
\item \noindent The example in Figure~\ref{fig:case_study}(d) shows the language comprehension ability of LTG in scene text words. The question requires the model to understand the linguistic attributes ``number''. Although the model locates the region which contain some 
wrong words, LTG can filter out the words that are not numbers with its natural language understanding ability.
\end{itemize}

\section{Conclusion}
We have presented the ``Locate Then Generate~(LTG)" framework that explicitly aligns 
the linguistic semantics and visual semantics of scene text through bounding boxes, which benefits the STVQA task. With the answer location module and the answer generation module, LTG boosts the non-pre-training baselines by $+6.06\%$ in absolute accuracy on the TextVQA challenge and outperforms the best non-pre-training 
 model by $+6.92\%$ on the ST-VQA challenge. 
It is noteworthy that LTG even outperforms TAP~\citep{yang2021tap} pre-trained on OCR-CC by $1.83\%$ on the ST-VQA dataset.  

\section{Acknowledgments}
This research was supported by the National Key Research and Development Plan of China (Grant No. 2022YFB3103100), the National Natural Science Foundation of China (Grant No. 62276245), and Anhui Provincial Natural Science Foundation (Grant No. 2008085J31).

\appendix

\bibliography{aaai23.bib}

\begin{thebibliography}{47}
\providecommand{\natexlab}[1]{#1}

\bibitem[{Alayrac et~al.(2022)Alayrac, Donahue, Luc, Miech, Barr, Hasson, Lenc,
  Mensch, Millican, Reynolds, Ring, Rutherford, Cabi, Han, Gong, Samangooei,
  Monteiro, Menick, Borgeaud, Brock, Nematzadeh, Sharifzadeh, Binkowski,
  Barreira, Vinyals, Zisserman, and Simonyan}]{Alayrac2022FlamingoAV}
Alayrac, J.-B.; Donahue, J.; Luc, P.; Miech, A.; Barr, I.; Hasson, Y.; Lenc,
  K.; Mensch, A.; Millican, K.; Reynolds, M.; Ring, R.; Rutherford, E.; Cabi,
  S.; Han, T.; Gong, Z.; Samangooei, S.; Monteiro, M.; Menick, J.; Borgeaud,
  S.; Brock, A.; Nematzadeh, A.; Sharifzadeh, S.; Binkowski, M.; Barreira, R.;
  Vinyals, O.; Zisserman, A.; and Simonyan, K. 2022.
\newblock Flamingo: a Visual Language Model for Few-Shot Learning.
\newblock \emph{ArXiv}, abs/2204.14198.

\bibitem[{Anderson et~al.(2018{\natexlab{a}})Anderson, He, Buehler, Teney,
  Johnson, Gould, and Zhang}]{anderson2018bottom}
Anderson, P.; He, X.; Buehler, C.; Teney, D.; Johnson, M.; Gould, S.; and
  Zhang, L. 2018{\natexlab{a}}.
\newblock Bottom-up and top-down attention for image captioning and visual
  question answering.
\newblock In \emph{Proceedings of the IEEE conference on computer vision and
  pattern recognition}, 6077--6086.

\bibitem[{Anderson et~al.(2018{\natexlab{b}})Anderson, Wu, Teney, Bruce,
  Johnson, S{\"u}nderhauf, Reid, Gould, and Van
  Den~Hengel}]{anderson2018vision}
Anderson, P.; Wu, Q.; Teney, D.; Bruce, J.; Johnson, M.; S{\"u}nderhauf, N.;
  Reid, I.; Gould, S.; and Van Den~Hengel, A. 2018{\natexlab{b}}.
\newblock Vision-and-language navigation: Interpreting visually-grounded
  navigation instructions in real environments.
\newblock In \emph{Proceedings of the IEEE conference on computer vision and
  pattern recognition}, 3674--3683.

\bibitem[{Antol et~al.(2015)Antol, Agrawal, Lu, Mitchell, Batra, Zitnick, and
  Parikh}]{antol2015vqa}
Antol, S.; Agrawal, A.; Lu, J.; Mitchell, M.; Batra, D.; Zitnick, C.~L.; and
  Parikh, D. 2015.
\newblock Vqa: Visual question answering.
\newblock In \emph{Proceedings of the IEEE international conference on computer
  vision}, 2425--2433.

\bibitem[{Bigham et~al.(2010)Bigham, Jayant, Ji, Little, Miller, Miller,
  Miller, Tatarowicz, White, White, and Yeh}]{bighamVizWiz}
Bigham, J.~P.; Jayant, C.; Ji, H.; Little, G.; Miller, A.; Miller, R.~C.;
  Miller, R.; Tatarowicz, A.; White, B.; White, S.; and Yeh, T. 2010.
\newblock VizWiz: Nearly Real-Time Answers to Visual Questions.
\newblock In \emph{Proceedings of the 23nd Annual ACM Symposium on User
  Interface Software and Technology}, UIST '10, 333–342. New York, NY, USA:
  Association for Computing Machinery.
\newblock ISBN 9781450302715.

\bibitem[{Biten et~al.(2022)Biten, Litman, Xie, Appalaraju, and
  Manmatha}]{biten2022latr}
Biten, A.~F.; Litman, R.; Xie, Y.; Appalaraju, S.; and Manmatha, R. 2022.
\newblock Latr: Layout-aware transformer for scene-text vqa.
\newblock In \emph{Proceedings of the IEEE/CVF Conference on Computer Vision
  and Pattern Recognition}, 16548--16558.

\bibitem[{Biten et~al.(2019)Biten, Tito, Mafla, Gomez, Rusinol, Valveny,
  Jawahar, and Karatzas}]{biten2019scene}
Biten, A.~F.; Tito, R.; Mafla, A.; Gomez, L.; Rusinol, M.; Valveny, E.;
  Jawahar, C.; and Karatzas, D. 2019.
\newblock Scene text visual question answering.
\newblock In \emph{Proceedings of the IEEE/CVF international conference on
  computer vision}, 4291--4301.

\bibitem[{Chen et~al.(2020)Chen, Li, Yu, Kholy, Ahmed, Gan, Cheng, and
  Liu}]{chen2019uniter}
Chen, Y.-C.; Li, L.; Yu, L.; Kholy, A.~E.; Ahmed, F.; Gan, Z.; Cheng, Y.; and
  Liu, J. 2020.
\newblock UNITER: UNiversal Image-TExt Representation Learning.
\newblock In \emph{European Conference on Computer Vision}.

\bibitem[{Deng et~al.(2021)Deng, Yang, Chen, Zhou, and
  Li}]{DBLP:conf/iccv/DengYCZL21}
Deng, J.; Yang, Z.; Chen, T.; Zhou, W.; and Li, H. 2021.
\newblock TransVG: End-to-End Visual Grounding with Transformers.
\newblock In \emph{2021 {IEEE/CVF} International Conference on Computer Vision,
  {ICCV} 2021, Montreal, QC, Canada, October 10-17, 2021}, 1749--1759. {IEEE}.

\bibitem[{Gao et~al.(2021)Gao, Zhu, Wang, Li, Liu, Van~den Hengel, and
  Wu}]{gao2021structured}
Gao, C.; Zhu, Q.; Wang, P.; Li, H.; Liu, Y.; Van~den Hengel, A.; and Wu, Q.
  2021.
\newblock Structured multimodal attentions for textvqa.
\newblock \emph{IEEE Transactions on Pattern Analysis and Machine
  Intelligence}.

\bibitem[{Gao et~al.(2020)Gao, Li, Wang, Shan, and Chen}]{gao2020multi}
Gao, D.; Li, K.; Wang, R.; Shan, S.; and Chen, X. 2020.
\newblock Multi-modal graph neural network for joint reasoning on vision and
  scene text.
\newblock In \emph{Proceedings of the IEEE/CVF conference on computer vision
  and pattern recognition}, 12746--12756.

\bibitem[{Goyal et~al.(2017)Goyal, Khot, Summers-Stay, Batra, and
  Parikh}]{goyal2017making}
Goyal, Y.; Khot, T.; Summers-Stay, D.; Batra, D.; and Parikh, D. 2017.
\newblock Making the v in vqa matter: Elevating the role of image understanding
  in visual question answering.
\newblock In \emph{Proceedings of the IEEE conference on computer vision and
  pattern recognition}, 6904--6913.

\bibitem[{Gurari et~al.(2018)Gurari, Li, Stangl, Guo, Lin, Grauman, Luo, and
  Bigham}]{gurari2018vizwiz}
Gurari, D.; Li, Q.; Stangl, A.~J.; Guo, A.; Lin, C.; Grauman, K.; Luo, J.; and
  Bigham, J.~P. 2018.
\newblock Vizwiz grand challenge: Answering visual questions from blind people.
\newblock In \emph{Proceedings of the IEEE conference on computer vision and
  pattern recognition}, 3608--3617.

\bibitem[{Han, Huang, and Han(2020)}]{han2020finding}
Han, W.; Huang, H.; and Han, T. 2020.
\newblock Finding the evidence: Localization-aware answer prediction for text
  visual question answering.
\newblock \emph{arXiv preprint arXiv:2010.02582}.

\bibitem[{Hu et~al.(2020)Hu, Singh, Darrell, and Rohrbach}]{hu2020iterative}
Hu, R.; Singh, A.; Darrell, T.; and Rohrbach, M. 2020.
\newblock Iterative answer prediction with pointer-augmented multimodal
  transformers for textvqa.
\newblock In \emph{Proceedings of the IEEE/CVF Conference on Computer Vision
  and Pattern Recognition}, 9992--10002.

\bibitem[{Jiang et~al.(2018)Jiang, Natarajan, Chen, Rohrbach, Batra, and
  Parikh}]{jiang2018pythia}
Jiang, Y.; Natarajan, V.; Chen, X.; Rohrbach, M.; Batra, D.; and Parikh, D.
  2018.
\newblock Pythia v0. 1: the winning entry to the vqa challenge 2018.
\newblock \emph{arXiv preprint arXiv:1807.09956}.

\bibitem[{Kamath et~al.(2021)Kamath, Singh, LeCun, Synnaeve, Misra, and
  Carion}]{KamathSLSMC21}
Kamath, A.; Singh, M.; LeCun, Y.; Synnaeve, G.; Misra, I.; and Carion, N. 2021.
\newblock {MDETR} - Modulated Detection for End-to-End Multi-Modal
  Understanding.
\newblock In \emph{2021 {IEEE/CVF} International Conference on Computer Vision,
  {ICCV} 2021, Montreal, QC, Canada, October 10-17, 2021}, 1760--1770. {IEEE}.

\bibitem[{Kant et~al.(2020)Kant, Batra, Anderson, Schwing, Parikh, Lu, and
  Agrawal}]{kant2020spatially}
Kant, Y.; Batra, D.; Anderson, P.; Schwing, A.; Parikh, D.; Lu, J.; and
  Agrawal, H. 2020.
\newblock Spatially aware multimodal transformers for textvqa.
\newblock In \emph{European Conference on Computer Vision}, 715--732. Springer.

\bibitem[{Karatzas et~al.(2015)Karatzas, Gomez-Bigorda, Nicolaou, Ghosh,
  Bagdanov, Iwamura, Matas, Neumann, Chandrasekhar, Lu
  et~al.}]{karatzas2015icdar}
Karatzas, D.; Gomez-Bigorda, L.; Nicolaou, A.; Ghosh, S.; Bagdanov, A.;
  Iwamura, M.; Matas, J.; Neumann, L.; Chandrasekhar, V.~R.; Lu, S.; et~al.
  2015.
\newblock ICDAR 2015 competition on robust reading.
\newblock In \emph{2015 13th international conference on document analysis and
  recognition (ICDAR)}, 1156--1160. IEEE.

\bibitem[{Kazemzadeh et~al.(2014)Kazemzadeh, Ordonez, Matten, and
  Berg}]{kazemzadeh2014referitgame}
Kazemzadeh, S.; Ordonez, V.; Matten, M.; and Berg, T. 2014.
\newblock Referitgame: Referring to objects in photographs of natural scenes.
\newblock In \emph{Proceedings of the 2014 conference on empirical methods in
  natural language processing (EMNLP)}, 787--798.

\bibitem[{Krishna et~al.(2017)Krishna, Zhu, Groth, Johnson, Hata, Kravitz,
  Chen, Kalantidis, Li, Shamma et~al.}]{krishna2017visual}
Krishna, R.; Zhu, Y.; Groth, O.; Johnson, J.; Hata, K.; Kravitz, J.; Chen, S.;
  Kalantidis, Y.; Li, L.-J.; Shamma, D.~A.; et~al. 2017.
\newblock Visual genome: Connecting language and vision using crowdsourced
  dense image annotations.
\newblock \emph{International journal of computer vision}, 123(1): 32--73.

\bibitem[{Kuznetsova et~al.(2020)Kuznetsova, Rom, Alldrin, Uijlings, Krasin,
  Pont-Tuset, Kamali, Popov, Malloci, Kolesnikov et~al.}]{kuznetsova2020open}
Kuznetsova, A.; Rom, H.; Alldrin, N.; Uijlings, J.; Krasin, I.; Pont-Tuset, J.;
  Kamali, S.; Popov, S.; Malloci, M.; Kolesnikov, A.; et~al. 2020.
\newblock The open images dataset v4.
\newblock \emph{International Journal of Computer Vision}, 128(7): 1956--1981.

\bibitem[{Lewis et~al.(2019)Lewis, Liu, Goyal, Ghazvininejad, Mohamed, Levy,
  Stoyanov, and Zettlemoyer}]{lewis2019bart}
Lewis, M.; Liu, Y.; Goyal, N.; Ghazvininejad, M.; Mohamed, A.; Levy, O.;
  Stoyanov, V.; and Zettlemoyer, L. 2019.
\newblock Bart: Denoising sequence-to-sequence pre-training for natural
  language generation, translation, and comprehension.
\newblock \emph{arXiv preprint arXiv:1910.13461}.

\bibitem[{Li et~al.(2020)Li, He, Zhou, Zhang, and Baldridge}]{li2020mapping}
Li, Y.; He, J.; Zhou, X.; Zhang, Y.; and Baldridge, J. 2020.
\newblock Mapping natural language instructions to mobile UI action sequences.
\newblock \emph{arXiv preprint arXiv:2005.03776}.

\bibitem[{Liu et~al.(2020{\natexlab{a}})Liu, Xu, Wu, Du, Jia, and
  Tan}]{liu2020cascade}
Liu, F.; Xu, G.; Wu, Q.; Du, Q.; Jia, W.; and Tan, M. 2020{\natexlab{a}}.
\newblock Cascade reasoning network for text-based visual question answering.
\newblock In \emph{Proceedings of the 28th ACM International Conference on
  Multimedia}, 4060--4069.

\bibitem[{Liu et~al.(2020{\natexlab{b}})Liu, Chen, Shen, He, Jin, and
  Wang}]{DBLP:conf/cvpr/LiuCSHJW20}
Liu, Y.; Chen, H.; Shen, C.; He, T.; Jin, L.; and Wang, L. 2020{\natexlab{b}}.
\newblock ABCNet: Real-Time Scene Text Spotting With Adaptive Bezier-Curve
  Network.
\newblock In \emph{2020 {IEEE/CVF} Conference on Computer Vision and Pattern
  Recognition, {CVPR} 2020, Seattle, WA, USA, June 13-19, 2020}, 9806--9815.
  Computer Vision Foundation / {IEEE}.

\bibitem[{Liu et~al.(2019)Liu, Ott, Goyal, Du, Joshi, Chen, Levy, Lewis,
  Zettlemoyer, and Stoyanov}]{liu2019roberta}
Liu, Y.; Ott, M.; Goyal, N.; Du, J.; Joshi, M.; Chen, D.; Levy, O.; Lewis, M.;
  Zettlemoyer, L.; and Stoyanov, V. 2019.
\newblock Roberta: A robustly optimized bert pretraining approach.
\newblock \emph{arXiv preprint arXiv:1907.11692}.

\bibitem[{Lu et~al.(2021)Lu, Fan, Wang, Oh, and Ros{\'e}}]{lu2021localize}
Lu, X.; Fan, Z.; Wang, Y.; Oh, J.; and Ros{\'e}, C.~P. 2021.
\newblock Localize, group, and select: Boosting text-vqa by scene text
  modeling.
\newblock In \emph{Proceedings of the IEEE/CVF International Conference on
  Computer Vision}, 2631--2639.

\bibitem[{Raffel et~al.(2020)Raffel, Shazeer, Roberts, Lee, Narang, Matena,
  Zhou, Li, Liu et~al.}]{raffel2020exploring}
Raffel, C.; Shazeer, N.; Roberts, A.; Lee, K.; Narang, S.; Matena, M.; Zhou,
  Y.; Li, W.; Liu, P.~J.; et~al. 2020.
\newblock Exploring the limits of transfer learning with a unified text-to-text
  transformer.
\newblock \emph{J. Mach. Learn. Res.}, 21(140): 1--67.

\bibitem[{Ren et~al.(2017)Ren, He, Girshick, and
  Sun}]{DBLP:journals/pami/RenHG017}
Ren, S.; He, K.; Girshick, R.~B.; and Sun, J. 2017.
\newblock Faster {R-CNN:} Towards Real-Time Object Detection with Region
  Proposal Networks.
\newblock \emph{{IEEE} Trans. Pattern Anal. Mach. Intell.}, 39(6): 1137--1149.

\bibitem[{Rezatofighi et~al.(2019)Rezatofighi, Tsoi, Gwak, Sadeghian, Reid, and
  Savarese}]{Rezatofighi_2018_CVPR}
Rezatofighi, H.; Tsoi, N.; Gwak, J.; Sadeghian, A.; Reid, I.; and Savarese, S.
  2019.
\newblock Generalized intersection over union: A metric and a loss for bounding
  box regression.
\newblock In \emph{Proceedings of the IEEE/CVF conference on computer vision
  and pattern recognition}, 658--666.

\bibitem[{Sidorov et~al.(2020)Sidorov, Hu, Rohrbach, and
  Singh}]{sidorov2020textcaps}
Sidorov, O.; Hu, R.; Rohrbach, M.; and Singh, A. 2020.
\newblock Textcaps: a dataset for image captioning with reading comprehension.
\newblock In \emph{European conference on computer vision}, 742--758. Springer.

\bibitem[{Singh et~al.(2019)Singh, Natarajan, Shah, Jiang, Chen, Batra, Parikh,
  and Rohrbach}]{singh2019towards}
Singh, A.; Natarajan, V.; Shah, M.; Jiang, Y.; Chen, X.; Batra, D.; Parikh, D.;
  and Rohrbach, M. 2019.
\newblock Towards vqa models that can read.
\newblock In \emph{Proceedings of the IEEE/CVF conference on computer vision
  and pattern recognition}, 8317--8326.

\bibitem[{Vaswani et~al.(2017)Vaswani, Shazeer, Parmar, Uszkoreit, Jones,
  Gomez, Kaiser, and Polosukhin}]{vaswani2017attention}
Vaswani, A.; Shazeer, N.; Parmar, N.; Uszkoreit, J.; Jones, L.; Gomez, A.~N.;
  Kaiser, {\L}.; and Polosukhin, I. 2017.
\newblock Attention is all you need.
\newblock \emph{Advances in neural information processing systems}, 30.

\bibitem[{Veit et~al.(2016)Veit, Matera, Neumann, Matas, and
  Belongie}]{veit2016coco}
Veit, A.; Matera, T.; Neumann, L.; Matas, J.; and Belongie, S. 2016.
\newblock Coco-text: Dataset and benchmark for text detection and recognition
  in natural images.
\newblock \emph{arXiv preprint arXiv:1601.07140}.

\bibitem[{Wang, Jin, and Ding(2022)}]{wang2022lilt}
Wang, J.; Jin, L.; and Ding, K. 2022.
\newblock Lilt: A simple yet effective language-independent layout transformer
  for structured document understanding.
\newblock \emph{arXiv preprint arXiv:2202.13669}.

\bibitem[{Wang et~al.(2022)Wang, Yang, Hu, Li, Lin, Gan, Liu, Liu, and
  Wang}]{wang2022git}
Wang, J.; Yang, Z.; Hu, X.; Li, L.; Lin, K.; Gan, Z.; Liu, Z.; Liu, C.; and
  Wang, L. 2022.
\newblock GIT: A Generative Image-to-text Transformer for Vision and Language.
\newblock \emph{arXiv preprint arXiv:2205.14100}.

\bibitem[{Wang et~al.(2021)Wang, Yu, Yu, Dai, Tsvetkov, and
  Cao}]{wang2021simvlm}
Wang, Z.; Yu, J.; Yu, A.~W.; Dai, Z.; Tsvetkov, Y.; and Cao, Y. 2021.
\newblock Simvlm: Simple visual language model pretraining with weak
  supervision.
\newblock \emph{arXiv preprint arXiv:2108.10904}.

\bibitem[{Xu et~al.(2020)Xu, Li, Cui, Huang, Wei, and Zhou}]{xu2020layoutlm}
Xu, Y.; Li, M.; Cui, L.; Huang, S.; Wei, F.; and Zhou, M. 2020.
\newblock Layoutlm: Pre-training of text and layout for document image
  understanding.
\newblock In \emph{Proceedings of the 26th ACM SIGKDD International Conference
  on Knowledge Discovery \& Data Mining}, 1192--1200.

\bibitem[{Yang et~al.(2019)Yang, Gong, Wang, Huang, Yu, and
  Luo}]{DBLP:conf/iccv/YangGWHYL19}
Yang, Z.; Gong, B.; Wang, L.; Huang, W.; Yu, D.; and Luo, J. 2019.
\newblock A Fast and Accurate One-Stage Approach to Visual Grounding.
\newblock In \emph{2019 {IEEE/CVF} International Conference on Computer Vision,
  {ICCV} 2019, Seoul, Korea (South), October 27 - November 2, 2019},
  4682--4692. {IEEE}.

\bibitem[{Yang et~al.(2021)Yang, Lu, Wang, Yin, Florencio, Wang, Zhang, Zhang,
  and Luo}]{yang2021tap}
Yang, Z.; Lu, Y.; Wang, J.; Yin, X.; Florencio, D.; Wang, L.; Zhang, C.; Zhang,
  L.; and Luo, J. 2021.
\newblock Tap: Text-aware pre-training for text-vqa and text-caption.
\newblock In \emph{Proceedings of the IEEE/CVF conference on computer vision
  and pattern recognition}, 8751--8761.

\bibitem[{Young et~al.(2014)Young, Lai, Hodosh, and
  Hockenmaier}]{young2014image}
Young, P.; Lai, A.; Hodosh, M.; and Hockenmaier, J. 2014.
\newblock From image descriptions to visual denotations: New similarity metrics
  for semantic inference over event descriptions.
\newblock \emph{Transactions of the Association for Computational Linguistics},
  2: 67--78.

\bibitem[{Yu et~al.(2018)Yu, Lin, Shen, Yang, Lu, Bansal, and Berg}]{YuMAttNet}
Yu, L.; Lin, Z.; Shen, X.; Yang, J.; Lu, X.; Bansal, M.; and Berg, T.~L. 2018.
\newblock MAttNet: Modular Attention Network for Referring Expression
  Comprehension.
\newblock \emph{CoRR}, abs/1801.08186.

\bibitem[{Zeng et~al.(2021)Zeng, Zhang, Zhou, and Yang}]{zeng2021beyond}
Zeng, G.; Zhang, Y.; Zhou, Y.; and Yang, X. 2021.
\newblock Beyond OCR+ VQA: involving OCR into the flow for robust and accurate
  TextVQA.
\newblock In \emph{Proceedings of the 29th ACM International Conference on
  Multimedia}, 376--385.

\bibitem[{Zhang et~al.(2021)Zhang, Ding, Peng, Fu, and Wang}]{9186709}
Zhang, C.; Ding, W.; Peng, G.; Fu, F.; and Wang, W. 2021.
\newblock Street View Text Recognition With Deep Learning for Urban Scene
  Understanding in Intelligent Transportation Systems.
\newblock \emph{IEEE Transactions on Intelligent Transportation Systems},
  22(7): 4727--4743.

\bibitem[{Zhang, Niu, and Chang(2018)}]{ZhangNC18}
Zhang, H.; Niu, Y.; and Chang, S. 2018.
\newblock Grounding Referring Expressions in Images by Variational Context.
\newblock In \emph{2018 {IEEE} Conference on Computer Vision and Pattern
  Recognition, {CVPR} 2018, Salt Lake City, UT, USA, June 18-22, 2018},
  4158--4166. Computer Vision Foundation / {IEEE} Computer Society.

\bibitem[{Zhu et~al.(2021)Zhu, Gao, Wang, and Wu}]{zhu2021simple}
Zhu, Q.; Gao, C.; Wang, P.; and Wu, Q. 2021.
\newblock Simple is not easy: A simple strong baseline for textvqa and
  textcaps.
\newblock In \emph{Proceedings of the AAAI Conference on Artificial
  Intelligence}, volume~35, 3608--3615.

\end{thebibliography}

\end{document}